\documentclass[accepted]{uai2023} 

\usepackage[american]{babel}

\usepackage[round]{natbib} 
\usepackage{mathtools} 
\usepackage{booktabs} 
\usepackage{tikz} 
\usepackage{xcolor}
\usepackage{url}            
\usepackage{nicefrac}       
\usepackage{microtype}      
\usepackage{natbib} 
\usepackage{xcolor}         
\usepackage{algorithm}
\usepackage[noend]{algpseudocode}
\usepackage{graphicx}
\usepackage{pifont}
\usepackage{adjustbox}
\usepackage{subfigure}
\usepackage{amsthm}
\usepackage{amssymb}
\usepackage{multirow}
\usepackage{makecell}

\usepackage{colortbl}
\usepackage{tabulary}
\usepackage{etoolbox}
\usepackage{pgf} 
\usepackage{collcell}

\definecolor{high}{HTML}{77b300}  
\definecolor{low}{HTML}{0066ff}  
\newcommand*{\opacity}{65}
\newcommand*{\minvalF}{0.64}
\newcommand*{\maxvalF}{0.94}
\newcommand*{\minvalC}{0.46}
\newcommand*{\maxvalC}{0.93}
\newcommand*{\minvalI}{0.38}
\newcommand*{\maxvalI}{0.69}
\newcommand{\gradientF}[1]{
    \ifdimcomp{#1pt}{>}{\maxvalF pt}{#1}{
    \ifdimcomp{#1pt}{<}{\minvalF pt}{#1}{
        \ifdimcomp{#1pt}{>}{0.92 pt}{} {
            \pgfmathparse{int(round(0.6*(100*(#1/(\maxvalF-\minvalF))-(\minvalF*(100/(\maxvalF-\minvalF))))))}
            \xdef\tempa{\pgfmathresult}
            \cellcolor{high!\tempa!low!\opacity} #1
        }
        \ifdimcomp{#1pt}{<}{0.93 pt}{} {
            \pgfmathparse{int(round(100))}
            \xdef\tempa{\pgfmathresult}
            \cellcolor{high!\tempa!low!\opacity} #1
        }
    }}
 }
 
 \newcommand{\gradientC}[1]{
    \ifdimcomp{#1pt}{>}{\maxvalC pt}{#1}{
    \ifdimcomp{#1pt}{<}{\minvalC pt}{#1}{
        \ifdimcomp{#1pt}{>}{0.91 pt}{} {
            \pgfmathparse{int(round(0.6*(100*(#1/(\maxvalC-\minvalC))-(\minvalC*(100/(\maxvalC-\minvalC))))))}
            \xdef\tempa{\pgfmathresult}
            \cellcolor{high!\tempa!low!\opacity} #1
        }
        \ifdimcomp{#1pt}{<}{0.92 pt}{} {
            \pgfmathparse{int(round(100))}
            \xdef\tempa{\pgfmathresult}
            \cellcolor{high!\tempa!low!\opacity} #1
        }
    }}
 }
 
 \newcommand{\gradientI}[1]{
    \ifdimcomp{#1pt}{>}{\maxvalI pt}{#1}{
    \ifdimcomp{#1pt}{<}{\minvalI pt}{#1}{
        \ifdimcomp{#1pt}{>}{0.67 pt}{} {
            \pgfmathparse{int(round(0.6*(100*(#1/(\maxvalI-\minvalI))-(\minvalI*(100/(\maxvalI-\minvalI))))))}
            \xdef\tempa{\pgfmathresult}
            \cellcolor{high!\tempa!low!\opacity} #1
        }
        \ifdimcomp{#1pt}{<}{0.68 pt}{} {
            \pgfmathparse{int(round(100))}
            \xdef\tempa{\pgfmathresult}
            \cellcolor{high!\tempa!low!\opacity} #1
        }
    }}
 }

\newcommand{\tabyes}{\ding{51}}%
\newcommand{\tabno}{\ding{55}}
\newtheorem{definition}{Definition}
\newtheorem{theorem}{Theorem}
\newcommand{\rom}[1]{\uppercase\expandafter{\romannumeral #1\relax}}

\title{FEATHERS: Federated Architecture and Hyperparameter Search}

\author[1]{\href{mailto:<jonas.seng@tu-darmstadt.de>}{Jonas Seng}{}}
\author[1]{Pooja Prasad}
\author[1, 2]{Devendra Singh Dhami}
\author[1, 2]{Martin Mundt}
\author[1, 2, 3]{Kristian Kersting}
\affil[1]{%
    Computer Science Department \\ AIML, TU Darmstadt \\ Germany
}
\affil[2]{%
    Hessian Center for AI (hessianAI)
}
\affil[3]{%
    Centre for Cognitive Science, TU Darmstadt
}
  
\begin{document}
\maketitle

\begin{abstract}

Deep neural architectures have profound impact on achieved performance in many of today's AI tasks, yet, their design still heavily relies on human prior knowledge and experience. 
Neural architecture search (NAS) together with hyperparameter optimization (HO) helps to reduce this dependence.
However, state of the art NAS and HO rapidly become infeasible with increasing amount of data being stored in a distributed fashion, typically  violating data privacy regulations such as GDPR and CCPA.
As a remedy, we introduce FEATHERS---\textbf{FE}derated \textbf{A}rchi\textbf{T}ecture and \textbf{H}yp\textbf{ER}parameter \textbf{S}earch,  a method that not only optimizes both neural architectures and optimization-related hyperparameters jointly in distributed data settings, but further adheres to data privacy  through the use of differential privacy (DP). 
We show that FEATHERS efficiently optimizes architectural and optimization-related hyperparameters alike, while 
demonstrating convergence on classification tasks at no detriment to model performance when complying with privacy constraints. 

\end{abstract}

\section{Introduction}
Federated learning (FL) is a distributed machine learning paradigm aiming to learn a shared model on data distributed at different locations without ever exchanging the data itself \citep{mcmahan2017communication}. It is a promising solution in several industries, such as finance or healthcare, where it is infeasible to share the data due to privacy and security regulations. 
\begin{table}[t]
    \centering
    \caption{\textbf{FEATHERS vs SOTA}. FEATHERS is the first method that enables both HO and NAS in a federated learning setting while providing privacy guarantees (DP).}
    \label{tab:method_comparison}
    \resizebox{\columnwidth}{!}{
        \begin{tabular}{l|c|c|c|c}
        Method & NAS & HO & DP & Fed. \\
        \hline
        DARTS\citep{liu2018darts} & \textcolor{green} \tabyes & \textcolor{red} \tabno & \textcolor{red} \tabno & \textcolor{red} \tabno\\
        DP-FNAS\citep{singh2020differentially} & \textcolor{green} \tabyes & \textcolor{red} \tabno & \textcolor{green} \tabyes & \textcolor{green} \tabyes\\
        FedNAS\citep{he2021fednas} & \textcolor{green} \tabyes & \textcolor{red} \tabno & \textcolor{red} \tabno & \textcolor{green} \tabyes\\
        DP-FTS-DE\citep{dai2021differentially} & \textcolor{red} \tabno & \textcolor{green} \tabyes & \textcolor{green} \tabyes & \textcolor{green} \tabyes\\
        FedEx\citep{khodak2021federated} & \textcolor{red} \tabno & \textcolor{green} \tabyes & \textcolor{red} \tabno & \textcolor{green} \tabyes\\
        \hline
        \textbf{FEATHERS} & \textcolor{green} \tabyes & \textcolor{green} \tabyes & \textcolor{green} \tabyes & \textcolor{green} \tabyes\\
    \end{tabular}
    }
\end{table}
As in classical machine learning (ML), neural architectures and optimization-related hyperparameters (hence simply referred to as hyperparameters) have to be selected in FL before training. Since even experts are likely to choose non-optimal architectures and hyperparameters, different neural architecture search (NAS) and hyperparameter optimization (HO) methods have been developed to automatically search for suitable architectures/hyperparameters \citep{kairouz2021advances, zoph2016neural,pham2018ENAS,liu2018darts,agrawal2021genetic}. With HO- and NAS-methods experts only have to define a search space over candidates instead of defining a specific rigid architecture and setting hyperparameters for a given ML-task. A search strategy is then applied to automatically find the optimal element within this space. 

To date, most NAS- and HO-methods are designed for classical ML-settings. As more and more data is being stored decentralized and privacy awareness is rising ~\citep{he2021fednas,khodak2021federated}, a number of approaches to perform NAS/HO in FL settings have lately been proposed. However, the latter still face a significant number of challenges. 
 First, current methods either optimize neural architectures \textit{or} hyperparameters; a serious obstacle as performing NAS and HO sequentially is costly, especially in FL settings where it is preferable to minimize the communication performed between devices. In addition, the choice of architectures and hyperparameters inherently depend on each other. For instance, adopting a deeper architecture may require selecting different learning rates in order to assure adequate update scaling, following gradient back-propagation through the network. Therefore architectures and hyperparameters should be optimized jointly.
 A second major challenge is that NAS- and HO-methods are traditionally not designed to be privacy-preserving. Throughout FL training, the server and clients exchange the updated parameters several times. In light of the growing concern about the disclosure of personal information\citep{fredrikson2015model_inversion} and the threat of adversarial attacks \citep{Ye2016MembershipInference}, both ML models and their training process, and hence NAS and HO approaches, should guarantee privacy in distributed settings. 

To address the above challenges, we propose a novel method: \textbf{FEATHERS}\footnote{We make our code publicly available at: {\footnotesize \url{https://anonymous.4open.science/r/FEATHERS-250B/}}.} - \textbf{FE}derated \textbf{A}rchi\textbf{T}ecture and \textbf{H}yp\textbf{ER}parameter \textbf{S}earch. As illustrated in Table \ref{tab:method_comparison}, FEATHERS is the first method to synergize architecture search and hyperparameter optimization, while enabling privacy preserving federated learning. 

\begin{figure}[t]
    \centering
        \includegraphics[width=1.35\columnwidth] {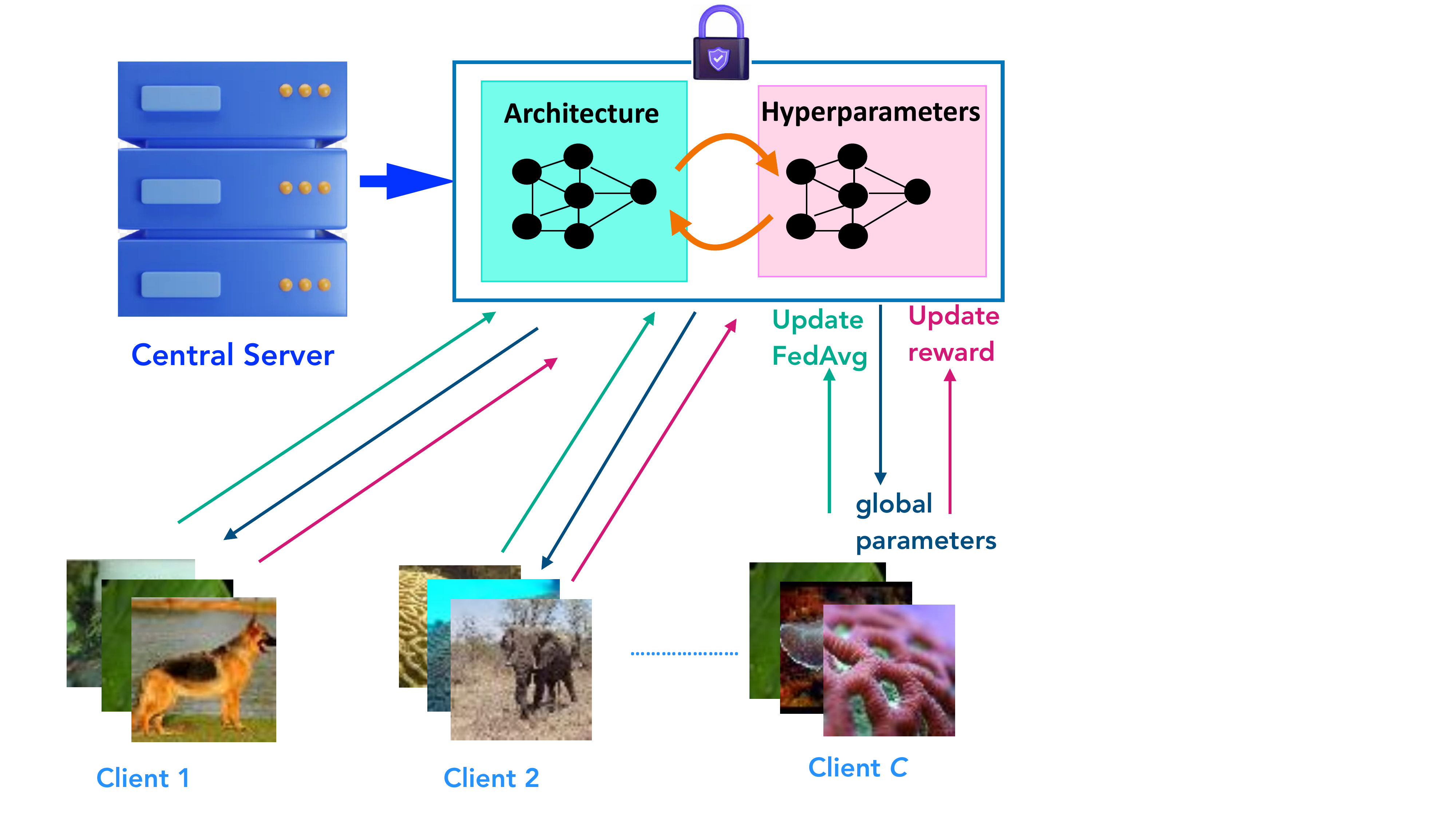}
        \caption{\textbf{FEATHERS Overview. } 
        FEATHERS jointly optimizes neural architectures and hyperparameters using data distributed across clients $\mathcal{C}$ while providing privacy-guarantee. In NAS-phase, the global architecture parameters and model parameters are computed using FedAvg over clients.The HO-phase selects hyperparameters for the subsequent NAS-phase based on the computed reward.}
    \label{fig:arch}
\end{figure}

The overall architecture of FEATHERS is shown in Fig.~\ref{fig:arch}. In essence, it consists of an HO- and a NAS-phase executed in an alternating fashion. In the HO phase, a $n$-bandit game is played to identify promising hyperparameters for a subsequent NAS-phase.
Then, a NAS-phase is performed for a multiple iterations using the identified hyperparameters. Differentiable, cell-baded, NAS \citep{liu2018darts} is used to allow "averaging`` architectures of several clients using FedAvg, thus obtaining a new global architecture. The global architecture obtained is then sent back to the clients and used for further training. The differentiability of the architecture entails a crucial property of FEATHERS: enabling a privacy preserving optimization-scheme using differential privacy (DP).

Overall, we make the following contributions:
\begin{enumerate}
    \item We propose a novel method, FEATHERS, that jointly optimizes neural architectures and hyperparameters in distributed data  settings.
    \item We prove that the HO phase of FEATHERS converges with high probability and that FEATHERS' convergence properties subsequently coincide with those of DARTS if the HO phase converges.
    \item By exploiting differentiability of model- and architectural parameters we provide privacy-guarantees during the search- and evaluation-stage via DP.
    \item We empirically show that FEATHERS converges towards well-suited architectures and hyperparameters on various classification tasks.
\end{enumerate}

We proceed as follows: After presenting related work we formally introduce the problem before moving to our proposed solution. A convergence analysis is then conducted, followed by our experimental evaluation of FEATHERS on several classification tasks. Finally, we conclude our work and show possible directions for future work.

\section{Related Work}

\textbf{Hyperparameter Optimization.}
Several works address hyperparameter optimization in federated learning \citep{koskela2018learning,mostafa2019robust}. Genetic CFL\citep{agrawal2021genetic} clusters edge devices based on the training hyperparameters and genetically modifies the parameters clusterwise. DP-FTS-DE\citep{dai2021differentially} integrates differential privacy into federated Thompson sampling with distributed exploration to preserve privacy and uses it for federated hyperparameter tuning. FLoRA\citep{zhou2021flora} uses a single-shot task by querying each client for several instantiations once and selects the best instantiations based on what the clients returned w.r.t. learning progress. FedEx\citep{khodak2021federated} uses a weight-sharing mechanism for hyperparameter optimization in the federated setting. 
In contrast, FEATHERS follows a few-shot policy as we adjust the hyperparameters several times during training.

\textbf{Neural Architecture Search.} NAS aims to automatically identify an optimal neural architecture for a given task. Most methods are based on reinforcement learning (RL) \citep{zoph2016neural,zoph2016transfer,zoph2018learning}, evolutionary algorithms (EAs) \citep{xie2017genetic,galvan2021neuroevolution,darwish2020survey} or gradient decent (GD) \citep{liu2018darts,dong2019searching,xie2018snas,li2020sgas}. Gradient based methods have been found to be more robust compared to the former \citep{zhu2021federated} and thus we adapt gradient-based neural architecture search in FEATHERS. Since this approach is differentiable it can be used in federated setting. Some recent NAS-methods for FL include Fed-NAS \citep{he2021fednas} that uses the gradient-based NAS method
MiLeNAS~\citep{he2020milenas} for personalized federated learning, and DP-FNAS~\citep{singh2020differentially}, which also adopted differentiable NAS~\citep{liu2018darts} combined with DP.

\textbf{Differential Privacy} DP has first been introduced in \citep{dwork2006DP} to protect private information in a dataset from queries based on arbitrary mechanisms. In recent years, select works have shown that ML-models, especially neural networks, carry private information of their training data in their parameters \citep{fredrikson2015model_inversion}. To protect such private data from leaking, DP has been successfully employed in various FL settings. In particular, it has been shown that SGD can be turned into a differentially private algorithm by simply adding an appropriate amount of noise to the parameters during training \citep{zhan2016DPSGD}.

\section{FEATHERS}
Our objective is to efficiently optimize neural architectures and hyperparameters in a joint manner in FL under privacy guarantees.
We now define our problem setup formally and present our proposed solution. 

\subsection{Problem Definition}
We consider a federated learning setting with a set of clients $\mathcal{C}$ of size $C$, each holding a dataset $D_1, ..., D_C$. The data of each client $c$ is split into training $\langle \mathbf{X}^{(c)}_{train}, \mathbf{y}^{(c)}_{train}\rangle$ and validation data $\langle \mathbf{X}^{(c)}_{val}, \mathbf{y}^{(c)}_{val}\rangle$ that is used to solve a supervised learning task. 
We aim to find an architecture $\mathbf{a} \in \mathcal{A}$ and hyperparameters $\mathbf{h} \in \mathcal{H}$ minimizing the global validation loss over all clients. Formally we phrase the problem as follows:
\begin{align}
    & \min_{\mathbf{a}, \mathbf{h}} \sum_{c \in \mathcal{C}} v_c \cdot \mathcal{L}_{\mathbf{a}, \mathbf{h}}(\mathbf{w^*}, \mathbf{X}_{val}^{(c)}, \mathbf{y}_{val}^{(c)}) \quad \text{with} \\ 
     \mathbf{w^*} = & \arg \min_{\mathbf{w}} \sum_{c \in \mathcal{C}} v_c \cdot \mathcal{L}_{\mathbf{a}, \mathbf{h}}(\mathbf{w}, \mathbf{X}_{train}^{(c)}, \mathbf{y}_{train}^{(c)})
\end{align}
where $\mathbf{w} \in \mathbb{R}^n$ refers to the parameters of a neural network (model parameters) and $v_c$ is the weight of a client $c$: 
\begin{equation}
    v_c := \frac{|\mathbf{X}^{(c)}_{val}|}{\sum_{c \in \mathcal{C}}|\mathbf{X}^{(c)}_{val}|}
\end{equation}
Note that in a FL settings, the global validation loss is a weighted sum of the client's local validation losses.
From now on, we denote the global training- and validation loss as $\mathcal{L}_{\mathbf{a}, \mathbf{h}}(\mathbf{w}, \mathbf{X}_{train}, \mathbf{y}_{train})$ and $\mathcal{L}_{\mathbf{a}, \mathbf{h}}(\mathbf{w}, \mathbf{X}_{val}, \mathbf{y}_{val})$ respectively.
Additionally, we require our method to be $\epsilon$-differential w.r.t. model parameters, architectural parameters and rewards.
For better readability, we omit DP in the definition of the optimization problem and will return to it in Section \ref{subsec:DP}.

\subsection{FEATHERS Architecture} \label{sec:FEATHERS}
FEATHERS operates in two stages, the \textit{search stage} and the \textit{evaluation stage}. The search stage consists of an alternating procedure: As a first step, an instantiation of hyperparameters is identified which is expected to achieve a high decrease in validation loss. In a second step, the identified instantiation is used to perform several optimization-steps of the architecture. The two steps are repeated until convergence.

In the evaluation stage the optimized architecture is retrained. Again, the HO-scheme from the search stage is applied to optimize hyperparameters.
We now describe the HO and NAS phase in detail.

\paragraph{Hyperparameter Optimization (HO).}
To identify well-working hyperparameters $\mathbf{h}$ we have to solve the following objective:
\begin{equation}
    \mathbf{h^*} = \arg \min_{\mathbf{h}} \mathcal{L}_{\mathbf{a^*}, \mathbf{h}}(\mathbf{w^*}; \mathbf{X}_{val}, \mathbf{y}_{val})
\end{equation}
Here, $\mathbf{w^*}$ denote model parameters minimizing the training-loss under architecture $\mathbf{a}^*$ and $\mathbf{a^*}$ are the architectural parameters minimizing the validation-loss under hyperparameters $\mathbf{h} \in \mathcal{H}$ where $\mathcal{H}$ is a discreete set of hyperparameter instantiations.
We solve the above by using a $n$-armed bandit-approach with a strategy similar to $\epsilon$-greedy as shown in Algorithm \ref{algo:FEATHERS_server}. \textbf{(Line 1-2)}: We start of by intializing the parameters, architecture and reward estimates. \textbf{(Line 6-12)}:
We then randomly sample $m$ hyperparameter instantiations from a distribution $\pi$ over $\mathcal{H}$. For each sampled instantiation one communication round of training is performed using the same weights $\mathbf{w}$ and architecture $\mathbf{a}$.. This yields an approximation of $\mathbf{a}^*$ and $\mathbf{w}^*$. Each client computes its local validation loss before and after performing local training using hyperparameters $\mathbf{h}$ and parameters $\mathbf{w}$ and $\mathbf{a}$. The losses are denoted as $\ell^{(c)}_{\mathbf{a}, \mathbf{w}}$ and $\ell^{(c)}_{\mathbf{a}', \mathbf{w}'}$ respectively. We compute the reward-signal $r_{\mathbf{h}}^{(e)}$ indicating how well instantiation $\mathbf{h}$ performed in HO-phase $e$ as:
\begin{equation}
    r_{\mathbf{h}}^{(e)} = \sum_{c \in \mathcal{C}} v_c \cdot \big(\ell^{(c)}_{\mathbf{a}, \mathbf{w}} - \ell^{(c)}_{\mathbf{a}', \mathbf{w}'}\big)
\end{equation}
In the above equation $v_c $ refers to the weight of each client. After testing each sampled $\mathbf{h}$ we obtain a vector $\mathbf{r}^{(e)}$ where each entry corresponds to one hyperparameter instantiations in $\mathcal{H}$: For instantiations $\mathbf{h}$ sampled in HO-round $e$, $\mathbf{r}^{(e)}$ contains the reward, all other entries are zero. The reward-estimates $\mathbf{r}$  are then updated using $\mathbf{r}^{(e)}$ by applying the update rule:
\begin{equation}
    \mathbf{r} = \mathbf{r} + (\mathbf{i} \circ \alpha \circ (\mathbf{r}^{(e)} - \mathbf{r})) + ((1 - \mathbf{i}) \circ (\alpha \circ \mathbf{r} -\mathbf{r})) 
\end{equation}
Here, $\circ$ is the Hadamard product, $\mathbf{i}$ is a binary vector indicating which hyperparameter instantiations were sampled in exploration round $e$ and $\alpha$ is a constant factor determining how aggressively the reward-estimate should be updated. If a hyperparameter instantiation was sampled at round $e$, its reward in $\mathbf{r}$ will be corrected by the error of the current reward estimate. All instantiations that were not sampled in $e$ get scaled down by $\alpha$ since well suited instantiations in an early stage of training might not be suitable in later stages anymore. For example, in the beginning of training, an instantiation with a higher learning rate might be more appropriate whereas in later training-stages lower learning rates should be chosen. 

The use of the reward estimates is three-fold: (1) The hyperparameter instantiation with the highest reward achieved so far will be used to train the supernet in the next NAS-phase. (2) Reward estimates determine the number of communication-rounds in the HO-phase before the HO-phase is starting: For this, we first compute the distribution $\pi = \text{softmax}(\mathbf{r})$ over instantiations using the reward estimates $\mathbf{r}$.
This allows to compute the entropy $H$:
\begin{equation}
    H = \sum_{\mathbf{h} \in \mathcal{H}} \text{ln}(\pi(\mathbf{h})) \cdot \pi(\mathbf{h})
\end{equation}
The number of HO-rounds performed next is determined by $\kappa = \text{rnd}(\beta H)$. Here, $\beta$ is a parameter to control the exploration-exploitation trade-off.
In the beginning, all rewards are set to 0, thus leading to a uniform distribution which has the maximum entropy. Over time the reward-estimates reflect which instantiations work well and which do not. Hence, $\pi$ gets less uniform and the entropy decreases over time, favoring exploitation over exploration in later training stages. (3) The distribution $\pi$ is used to sample hyperparameter-instantiations tested in the next HO-round.

\begin{algorithm}[t]
\caption{FEATHERS method server side}
\label{algo:FEATHERS_server}
\begin{algorithmic}[1]
  \Require set of clients $\mathcal{C}$, client weight $v_c \forall c \in \mathcal{C}$
  \Require search spaces $\mathcal{H}$, $\mathcal{A}$
  \State initialize parameters $\mathbf{w}$ and architecture $\mathbf{a}$
  \State initialize reward estimates $\mathbf{r} \gets \mathbf{0}$
  \State $\mathcal{P} \gets$ softmax($\mathbf{r}$)
  \For{$p$ in phases}
    \If{$p$ == 'ho'}:
        \State sample $n$ hyperparameters $\mathbf{h}$ from $\mathcal{P}$
        \State $\mathbf{r}_p \gets \mathbf{0}$
        \For{$h$ in $\mathbf{h}$}
            \State $l_1$, $l_2$, $\mathbf{w}^*$, $\mathbf{a}^* \gets$ client\_step($h$, $\mathbf{w}$, $\mathbf{a}$)
            \State $\mathbf{r}_p[h] \gets \sum_{c \in \mathcal{C}} v_c \cdot (l_1^{(c)} - l_c^{(c)})$ 
        \EndFor
        \State $\mathbf{r} \gets$ update\_rewards($\mathbf{r}$, $\mathbf{r}_p$)
        \State $\mathcal{P} \gets$ softmax($\mathbf{r})$
    \EndIf
    \If{$p$ == 'nas'}:
        \State $\mathbf{h}^* \gets \mathcal{H}[\arg \max_\mathbf{h} \mathbf{r}[\mathbf{h}]]$ 
        \For{$j$ in nas\_steps}
            \State $\mathbf{w}$, $\mathbf{a} \gets$ client\_steps($\mathbf{h}^*$, $\mathbf{w}$, $\mathbf{a}$)
        \EndFor
    \EndIf
  \EndFor
\end{algorithmic}
\end{algorithm}

\paragraph{Neural Architecture Search.}
Once the HO-phase yields a hyperparameter instantiation $\mathbf{h}$, the architecture is optimized under $\mathbf{h}$ for a certain number of communication rounds as shown in Algorithm \ref{algo:FEATHERS_server} \textbf{(Line 13-16)}, thereby solving:
\begin{align}\label{eq:DARTS}
    & \min_{\mathbf{a}} \mathcal{L}_{\mathbf{a}, \mathbf{h}}(\mathbf{w}^*, \mathbf{X}_{val}, \mathbf{y}_{val}) \\
    \text{where } \mathbf{w}^* = & \arg \min_{\mathbf{w}} \mathcal{L}_{\mathbf{a}, \mathbf{h}}(\mathbf{w}, \mathbf{X}_{train}, \mathbf{y}_{train})
\end{align}
Inspired by Differentiable Architecture Search (DARTS) \citep{liu2018darts} we solve this optimization problem as follows: We define our search space to be a space over \textit{cells}. A cell is a Directed Acyclic Graph (DAG) in which each node is a feature representation and each edge is a \textit{mixed operation}. The feature representation of some node $z$ is computed using all its parent-nodes and the mixed operations defining the edges between $z$ and its parent, i.e. for some node $z_j$ the representation is computed as:
$
    z_j = \sum_{i < j} o^{(z_i, z_j)}(\mathbf{x}^{z_i})
$
Here, $o^{(z_i, z_j)}$ is a mixed operation and $\mathbf{x}^{z_i}$ is the feature representation of node $z_i$.
A mixed operation connecting nodes $z_1$ and $z_2$ is defined as a weighted sum over a set of operations $\mathcal{O}$:
\begin{equation}
    o^{(z_1, z_2)} = \sum_{o \in \mathcal{O}} \frac{\exp(a_o^{(z_1, z_2)})}{\sum_{o' \in \mathcal{O}} \exp(a_{o'}^{(z_1, z_2)})}o(\mathbf{x})
\end{equation}
Here, $a_o^{(z_i, z_j)}$ are the architectural parameters to be learned. Since they fully describe the architecture, we will refer to them as the architecture $\mathbf{a}$ from now on.
Objective \ref{eq:DARTS} is solved by an alternating optimization of the architecture and model parameters. First, the architecture is updated by following the gradient ${\nabla_{\mathbf{a}} \mathcal{L}_{\mathbf{a}, \mathbf{h}}(\hat{\mathbf{w}}, \mathbf{X}_{val}, \mathbf{y}_{val})}$ where ${\hat{\mathbf{w}} = \mathbf{w} - \eta \nabla_{\mathbf{w}}\mathcal{L}_{\mathbf{a}, \mathbf{h}}(\mathbf{w}, \mathbf{X}_{train}, \mathbf{y}_{train})}$. As a second step the model parameters are updated by following the gradient ${\nabla_{\mathbf{w}}\mathcal{L}_{\mathbf{a}, \mathbf{h}}(\mathbf{w}, \mathbf{X}_{train}, \mathbf{y}_{train})}$.
Parameter-updates are performed on client-side in each communication round and yield new architectural and model parameters $\mathbf{a}'_c$ and $\mathbf{w}'_c$ for each client $c$ respectively. Since both, the architectural and model parameters, are parameters of a non-convex optimization problem with a differentiable loss-function, we use FedAvg to aggregate the model- and architecture parameters of all clients after each communication round.
We use two types of cells: Normal cells and reduction cells. Normal cells keep the dimensions of the input while reduction cells apply an additional reduction-operation.

\textbf{Discretizing the Architecture.}
Since differentiable NAS requires a continuous relaxation of the architectural sapce $\mathcal{A}$, the architecture learned by FEATHERS has to be discretized after training. This is done by selecting the top $k$ operations with the highest architectural weight over all cells. Also, no operation is allowed to connect the same two nodes.

\subsection{Differential Privacy}
\label{subsec:DP}
Although in FL no data is exchanged between server and clients, the parameters sent to the server still leak private information \citep{fredrikson2015model_inversion}. It has been shown that differential privacy (DP) can be used to provably protect private information encoded in these parameters during Stochastic Gradient Descent (SGD) \citep{zhan2016DPSGD}. We adapt this notion to both, model parameters and architectural parameters since both inherently depend on the data the model is trained on.
DP was introduced in \cite{dwork2006DP} and is defined as follows:
\begin{definition}
For any two datasets $D$, $D'$ that differ in exactly one record a mechanism $M$ is called $\epsilon$-differential private if $\forall x: \text{Pr}[M(D) = x] \leq \exp(\epsilon) \text{Pr}[M(D') = x]$ holds where $\text{Pr}[M(D) = x]$ refers to the probability of mechanism $M$ outputting $x$ if executed on $D$.
\end{definition}
In our case $M$ is the learning procedure, i.e. SGD. Making SGD differential private can be achieved by clipping gradients and adding Gaussian noise to the gradient of each sample w.r.t. the parameters, resulting in an algorithm called DP-SGD \citep{zhan2016DPSGD}. Updating model- and architectural parameters with DP-guarantees then becomes:
\begin{equation}
    \theta \leftarrow \alpha_{\theta} \frac{1}{B}\sum_{i=1}^B \nabla_{\theta} \mathcal{L}_{\mathbf{a}}(\mathbf{w}, \mathbf{x}^{(i)}) + \mathcal{N}(0, \sigma_{\theta}^2C_{\theta}^2\mathbf{I}) 
\end{equation}
In the above equation $\theta \in \{\mathbf{w}, \mathbf{a}\}$ refers to the model- or architectural parameters, $B$ is the batch-size, $\mathcal{N}$ is the normal distribution, $\sigma_{\theta}$ is a scaling-parameter, $C_{\theta}$ is the maximum gradient norm, $\mathbf{I}$ the identity matrix and $\alpha_{\theta}$ is the learning rate for parameters $\theta$. 
We use DP-SGD for learning both, the model parameters and the architecture. Hence, our method enjoys all convergence- and privacy-guarantees given by DP-SGD which can be controlled via the parameter $\epsilon$ \citep{zhan2016DPSGD}. As $\epsilon$ inversely depends on noise-parameters $\sigma$, for high $\epsilon$-values DP-SGD achieves approximately SGD-convergence while losing privacy-guarantees. For low $\epsilon$ we obtain strong privacy guarantees while losing convergence-guarantees.
It should be noted that FedAvg averages the parameters that have been computed by the clients. Since DP is closed under arbitrary post-processing, averaging does not break DP \citep{dwork2014DP}. Similarly, we apply DP on rewards sent to the server to "hide`` possible private information from data by adding small Gaussian noise with zero mean to the rewards. For the DP-variant of FEATHERS, see Appendix A.

\subsection{Convergence Analysis}
We now show that FEATHERS' convergence properties in distributed settings coincide with the convergence properties of DARTS in centralized settings with high probability, only scaled by a controllable factor arising from using FedAvg. For simplicity we do not consider adding DP in our analysis.
\begin{theorem}
Given a joint distribution $p(X_1, \dots, X_n, y)$ over random variables $X_1, \dots, X_n, y$ from which each client $c \in \mathcal{C}$ from a set of clients $\mathcal{C}$ samples a dataset $\langle \mathbf{X}^{(c)}, \mathbf{y}^{(c)}\rangle$, FEATHERS enjoys the same convergence properties as DARTS in a centralized setting if applied on a dataset $\langle \mathbf{X}, \mathbf{y}\rangle$ where $\mathbf{X} = \bigcup_{c \in \mathcal{C}} \mathbf{X}^{(c)}$ and $\mathbf{y} = \bigcup_{c \in \mathcal{C}} \mathbf{y}^{(c)}$.
\end{theorem}

\begin{proof}
We treat the HO-phase of FEATHERS as an oracle and assume that it returns well-suited hyperparameters $\mathbf{h}$. Once $\mathbf{h}$ was obtained, it is fixed for a certain number of communication rounds $\kappa$. In each communication round $i$ epochs of DARTS are performed locally on each client. Since we employ FedAvg to average model parameters after $i$ local epochs, we exploit that FedAvg converges  with rate $\mathcal{O}(\frac{1}{i \kappa})$ \citep{li2019FedAvgConvergence}. Since FedAvg converges and parameter-updates are only propagated during NAS-phases, it follows that FEATHERS enjoys the same convergence properties as DARTS in each NAS-phase scaled by the convergence of FedAvg $\mathcal{O}(\frac{1}{i \kappa})$.
\end{proof}
Since the above proof assumes that our method selects well-suited hyperparameters $\mathbf{h}$ for each NAS-phase, we will now show that the HO-phase converges with high probability in non-stationary bandit-environments.
\begin{theorem}\label{theo:rewards}
Given a fixed hyperparameter-space $\mathcal{H}$ and noisy, non-stationary rewards $r_{\mathbf{h}}^{(j)} \sim \mathcal{N}(\mu_{\mathbf{h}}^{(j)}, \sigma_{\mathbf{h}})$ where $\mu_{\mathbf{h}}^{(j)}$ is the expected value of the reward at iteration $j$, $\sigma_{\mathbf{h}}$ its standard deviation and $\mathbf{h} \in \mathcal{H}$, the HO-strategy of FEATHERS is at most off by $\alpha \cdot 3\sigma_{\mathbf{h}}$ for learning rate $\alpha$ with probability $0.997$ once $\mathbf{h} \in \mathcal{H}$ is sampled.
\end{theorem}
\begin{proof}
Our proof is inspired by convergence results for $\epsilon$-greedy strategies as stated in \citep{Sutton1998RL}. We assume that $| \mu_{\mathbf{h}}^{(j+1)} - \mu_{\mathbf{h}}^{(j)} | \leq \delta$ for all iterations and that we set $0 < \alpha < 1$ in the update rule. Since the softmax-function cannot evaluate to a point-mass, for the probability of each hyperparameter instantiation $\pi_i[\mathbf{h}] > 0$ holds. Thus, with $j$ approaching infinity, each $\mathbf{h} \in \mathcal{H}$ will be sampled infinitely many times, i.e. each instantiation will be sampled. At an iteration $j$, in the most extreme case, a certain $\mathbf{h} \in \mathcal{H}$ has not been sampled yet. Assume it gets sampled in iteration $j$. Since $r_{\mathbf{h}}^{(j)} \sim \mathcal{N}(\mu_{\mathbf{h}}^{(j)}, \sigma_{\mathbf{h}})$ and the current estimate reward-estimate $\mathbf{r}_{\mathbf{h}} = 0$, the update rule reads: $\mathbf{r}_{\mathbf{h}} = \alpha \cdot r_{\mathbf{h}}^{(j)}$. Since we assume all rewards being Gaussian distributed, the probability of obtaining a reward $r_{\mathbf{h}}^{(j)}$ in the range of $3 \sigma_{\mathbf{h}}$ is $0.997$. Since $0 < \alpha < 1$ holds, our estimate is at most $\pm \alpha \cdot 3 \sigma_{\mathbf{h}}$ of w.r.t. $\mu_{\mathbf{h}}^{(j)}$ in $99.7\%$ of the cases.
\end{proof}
As the above only considers the case in which our algorithm terminates after some $\mathbf{h} \in \mathcal{H}$ is sampled, we also have to consider the following case: Assume $\mathbf{h}$ is sampled at iteration $j$ and a reward-estimate is obtained. After that, $\mathbf{h}$ is not sampled for $k$ subsequent iterations. The following theorem gives bounds for how much off our estimate will be in this case.

\begin{theorem}\label{theo:reward_bound}
Under the assumptions of Theorem \ref{theo:rewards}, the reward estimate $r_{\mathbf{h}}^{(j+k)}$ will be at most off by $\alpha^k r_{\mathbf{h}}^{(j)} - (k \delta + \mu_{\mathbf{h}}^{(j)})$ assuming that $\mathbf{h}$ is sampled at iteration $j$ and not sampled for $k$ subsequent iterations.
\end{theorem}
\begin{proof}
By assumptions from Theorem \ref{theo:rewards}, the mean will be shifted by at most $k \delta$ after $k$ steps. Since the update rule for $r_{\mathbf{h}}$ is defined as $r_{\mathbf{h}} = \alpha r_{\mathbf{h}}$, the reward estimate after $k$ iterations in which $\mathbf{h}$ is not sampled is $\alpha^k r_{\mathbf{h}}^{(j)}$. It follows that, $k$ iterations after $\mathbf{h}$ was sampled, the reward estimate is off by at most $\alpha^k r_{\mathbf{h}}^{(j)} - (k \delta + \mu_{\mathbf{h}}^{(j)})$.
\end{proof}
It turns out that the above bound can be controlled by setting $\alpha \leq (1 + \frac{k \delta}{\mu^{(j)}})^{\frac{1}{k}}$ assuming we have access to $\mu^{(j)}$ (see Appendix F). In the case $\mu^{(j+1)} - \mu^{(j)} = \delta$, this relation guarantees that our reward estimate of some $\mathbf{h}$ is still optimal if $\mathbf{h}$ was not sampled for $k$ HO-rounds. Since we can assume that the loss decreases between HO-rounds, i.e. $\mu^{(j+1)} - \mu^{(j)} < 0$, the assumption $0 < \alpha < 1$ used in the above theorems is not violated. Using Theorem \ref{theo:rewards} we can assume that we have an estimate of $\mu^{(j)}$ fulfilling at least $\mu^{(j)} \pm \alpha \cdot 3\sigma_{\mathbf{h}}$ with high probability for some $\mathbf{h}$ sampled the first time in round $j$. Hence, the errors of reward-estimates can be controlled within reasonable bound given by Theorem \ref{theo:reward_bound} in subsequent rounds.

\begin{table*}[t!]
\centering
\caption{ \textbf{FEATHERS outperforms state of the art federated NAS- and HO-methods while being more flexible}. DARTS, FedEx and FEATHERS were compared in different federated learning settings as described in Section \ref{sec:exp_prot}. Each experiment was performed $5$ times and the mean accuracy and standard deviation are reported. Colors are interpolated from green to blue (high accuracy to low accuracy).}
\begin{tabular}{l|cc|cc|cc} 
Dataset                                     & \multicolumn{2}{c|}{Fashion-MNIST}                         & \multicolumn{2}{c|}{CIFAR-10}                               & \multicolumn{2}{c}{Tiny-Imagenet}                                      \\ 
                                            & \multicolumn{1}{c}{i.i.d} & non-i.i.d.          & \multicolumn{1}{c}{ i.i.d.} & non-i.i.d.          & \multicolumn{1}{c}{i.i.d.} & \multicolumn{1}{c}{non-i.id.}  \\ 
\midrule
DARTS (f, 5 clients)$^{\dag}$                        & $\gradientF{0.92} \pm 0.02$                 & $\gradientF{0.93} \pm 0.01$ & $\gradientC{0.92} \pm 0.02$         & $\gradientC{0.90} \pm 0.02$          & $\gradientI{0.67} \pm 0.02$                                 & $\gradientI{0.67} \pm 0.02$                                    \\
DARTS (f, 10 clients)$^{\dag}$                       & $\gradientF{0.91} \pm 0.02$                 & $\gradientF{0.92} \pm 0.02$          & $\gradientC{0.91} \pm 0.03$                  & $\gradientC{0.89} \pm 0.04$          &  $\gradientI{0.68} \pm 0.02$                                & $\gradientI{0.67} \pm 0.03$                                     \\
\multicolumn{1}{l|}{DARTS (f, 100 clients)$^{\dag}$} & $\gradientF{0.92} \pm 0.03$                                & $\gradientF{0.91} \pm 0.03$                         & $\gradientC{0.91} \pm 0.02$                                 & $\gradientC{0.89} \pm 0.03$                         & $\gradientI{0.67} \pm 0.02$                                  & $\gradientI{0.67} \pm 0.03$                                     \\
FedEx (5 clients)*                          & $\gradientF{0.82} \pm 0.01$                 & $\gradientF{0.81} \pm 0.01$          & $\gradientC{0.53} \pm 0.02$                  & $\gradientC{0.54} \pm 0.03$          & $\gradientI{0.43} \pm 0.04$                                 & $\gradientI{0.41} \pm 0.04$                                     \\
FedEx (10 clients)*                         & $\gradientF{0.78} \pm 0.03$                 & $\gradientF{0.78} \pm 0.02$          & $\gradientC{0.51} \pm 0.04$                  & $\gradientC{0.51} \pm 0.03$          & $\gradientI{0.41} \pm 0.03$                                 & $\gradientI{0.40} \pm 0.04$                                     \\
\multicolumn{1}{l|}{FedEx (100 clients)*}   & $\gradientF{0.65} \pm 0.03$                                &  $\gradientF{0.64} \pm 0.04$                        & $\gradientC{0.46} \pm 0.05$                                 & $\gradientC{0.47} \pm 0.04$                         & $\gradientI{0.38} \pm 0.04$                                  & $\gradientI{0.38} \pm 0.05$                                     \\ 
\midrule
FEATHERS (5 clients)                        & $\gradientF{0.94} \pm 0.01$        & $\gradientF{0.93} \pm 0.02$ & $\gradientC{0.93} \pm 0.03$         & $\gradientC{0.91} \pm 0.03$ & $\gradientI{0.69} \pm 0.02$                                  & $\gradientI{0.69} \pm 0.03$                                     \\
FEATHERS (10 clients)                       & $\gradientF{0.93} \pm 0.01$                 & $\gradientF{0.93} \pm 0.03$ & $\gradientC{0.92} \pm 0.02$         & $\gradientC{0.89} \pm 0.04$          & $\gradientI{0.68} \pm 0.02$                                  & $\gradientI{0.68} \pm 0.02$                                     \\
\multicolumn{1}{l|}{FEATHERS (100 clients)} &  $\gradientF{0.94} \pm 0.02$                               & $\gradientF{0.93} \pm 0.03$                         & $\gradientC{0.90} \pm 0.03$                                  & $\gradientC{0.89} \pm 0.03$                          & $\gradientI{0.68} \pm 0.03$                                 & $\gradientI{0.67} \pm 0.03$                                 \\
\multicolumn{7}{l}{\begin{tabular}[c]{@{}l@{}}*Training performed using architecture found by DARTS.\\ \dag The same hyperparameter-settings as described in \citep{liu2018darts} were used. \\ \end{tabular}} \\
\end{tabular}
\label{tab:results}
\end{table*}

\section{Experiments and Results} \label{sec:exp_prot}
In order to empirically demonstrate that FEATHERS is capable of jointly optimizing neural architectures and hyperparameters in FL settings with privacy guarantees, we investigate the following three questions: 
\begin{enumerate}
    \item[\textbf{Q1.}] Can FEATHERS compete with state of the art HO- and NAS-methods in FL settings at various scales and label skews (referred to as non-i.i.d.)?
    \item[\textbf{Q2.}] Does FEATHERS adjust the choices of instantiations over time to account for dynamics of training process?
    \item[\textbf{Q3.}] How well does FEATHERS perform if DP is employed to preserve privacy with respect to privacy-budget $\epsilon$?
\end{enumerate}

We next describe our experiment protocol including the employed datasets before presenting our results in detail.

\subsection{Experimental Protocol}
In our experiments, we analyzed FEATHERS on three image classification tasks: Fashion-MNIST which contains black-white images of 10 different articles of clothing to be categorized as well as CIFAR-10 and Tiny-Imagenet which contain colored images of 10/200 different categories respectively. The fourth task is a binary classification problem on a fraud detection dataset which contains anonymized bank-account data from bank-customers based on which the fraud risk (high or low) has to be predicted (see Appendix B). All datasets were partitioned randomly on a set of clients such that each client holds the approximately same number of samples. Since in FL it is common to have data unequally distributed across clients, we also conducted experiments in which we introduced a label skew in the data (referred to as \textit{non-i.i.d.} subsequently) .
We first executed the search stage of FEATHERS using a search space over CNN/MLP-architectures. For the evaluation-stage we used the best normal cell and reduction cell obtained in the search stage to build up a larger network (validation networks). For discretizing the architecture $k = 2$ was chosen in order to be comparable to other cell-based NAS-methods. We  then retrained the validation network and  optimized hyperparameters using the same HO-strategy as in the search stage. The results of the validation network were then reported. Since we assume a cross-silo setting, we allowed all clients to participate in each communication round. Additionally we tested both FEATHERS with and without DP for fraud detection, to show that adding DP does not prevent learning a suitable architecture.
Appendix C contains a detailed description of our search space.

We implemented FEATHERS in Python based on the flower framework for federated learning. All models were built using PyTorch and the clients were distributed on Nvidia DGX-clusters with A-100 40GB GPUs. Also the server was deployed on the same cluster, however, using a separate GPU to simulate a cross-silo federated learning setting with parallel client-execution.

\subsection{Results}
We are now ready to answer the posed research questions and will elaborate on each of them in more detail.
\paragraph{(Q1) FEATHERS achieves SOTA, independently of scale and label skew.}
First, we show that FEATHERS, while performing both HO and NAS, achieves state of the art results on Fashion-MNIST, CIFAR-10 and Tiny-Imagenet in Table \ref{tab:results}. Despite DARTS being based on hyperparameters that have traditionally manually been tuned for best results by humans, our method beats DARTS ($94\%$ vs. $92\%$ on Fashion-MNIST, $93\%$ vs. $92\%$ on CIFAR-10, $0.69\%$ vs. $0.67\%$ on Tiny-Imagenet) in most distributed learning settings while optimizing for a larger set of parameters. The same holds for label-skew scenarios. Introducing label skews does not seem to adversely affect its performance. However, if label-skew is present, a slight increase in the variance of our results can be seen. 

Second, to assess scalability of FEATHERS, we consider variance in results for increasing number of clients (marked in the rows of Table \ref{tab:results}).
We find that, in contrast to FedEx, the number of participating clients does not seem to have a negative influence on the performance of FEATHERS.
We hypothesize that FEATHER's stability is due to more extensive exploration. For an increasing number clients, each client holds a smaller subset of data since the datasets used have fixed size. Thus the stochastic gradients per client tend to have a higher variance, which in turn leads to higher variances in parameter-changes across clients. FEATHERS tests a set of hyperparameter instantiations on a frozen model before applying them for parameter-updates, whereas FedEx directly applies the chosen instantiations. Consequently, higher variance of gradients and FedEx' higher risk of choosing inappropriate hyperparameters can lead to poorly performing models. In contrast, FEATHERS tends to choose ''safer`` instantiations.

\begin{figure*}[t]
    \centering
        \includegraphics[width=\textwidth]{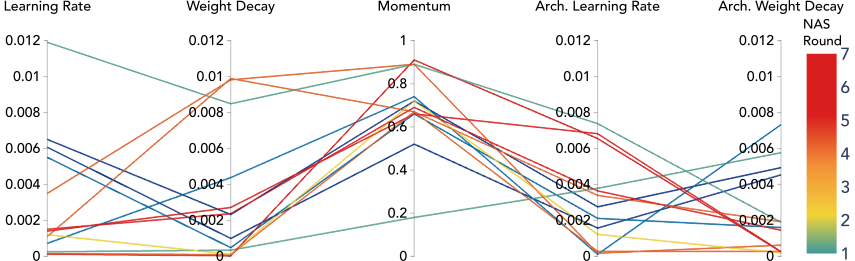}
        \caption{\textbf{FEATHERS adjusts hyperparameters over time.} The choices of hyperparameters are adapted during training to optimize the validation loss. In earlier stages (blue lines) higher learning rates are chosen whereas in later stages of training (red lines) lower learning rates are chosen. The figure shows hyperparameter-selections of three FEATHERS-runs on CIFAR-10. (Best viewed in color)}
    \label{fig:hype_conf}
\end{figure*}

Finally, in terms of runtime, FEATHERS ($\sim 2.5$ GPU-days) does not add significant overhead compared to DARTS ($\sim 2$ GPU-days). The additional HO-phase during the search stage adds an overhead of approximately 0.1-0.8 GPU-days, depending on the number of instantiations tested in each HO-round. In contrast, FedEx' runtime ($\sim 1$ GPU-day) is much lower compared to DARTS and FEATHERS since FedEx does not perform NAS and that it performs less exploration than our method. See Appendix G for a detailed listing of runtimes w.r.t. datasets and number of clients.
\paragraph{(Q2) FEATHERS dynamically adjusts hyperparameters.}
Figure \ref{fig:hype_conf} shows the hyperparameters selected by FEATHERS over time for three runs on CIFAR-10. We observe that our method chooses more ''cautious`` hyperparameters than engineers usually do. For example, in DARTS it is common to start with a learning rate of $0.025$, FEATHERS however chooses much lower learning rates most of the time.
Presumably this is due to the properties of our HO-algorithm: In the first HO-round it samples and tests a small subset of instantiations from $\mathcal{H}$ before greedily selecting the one leading to the highest decrease in validation loss. In this concrete example, this choice might lead FEATHERS to choose a lower learning rate than $0.025$, simply because there was no better sample.
In subsequent HO-rounds the goal is to learn a distribution over instantiations maximizing the reward in the long run. As SGD never truly converges due to its inherent stochasticity, a smaller learning rate is ultimately beneficial in the later stages of training, in order to avoid heavily perturbing away from a minimum (i.e. too large learning rates will ''overshoot``). 

Consequently, FEATHER's ''cautios`` instantiations entail more stable convergence. 
In that sense, FEATHERS mimics an annealing mechanism in later training stages, which find frequent use in Deep Learning problems. This observation further supports our claim that our method adjusts hyperparameters appropriately over time.

\begin{figure}[t]
    \centering
    \includegraphics[width=\columnwidth]{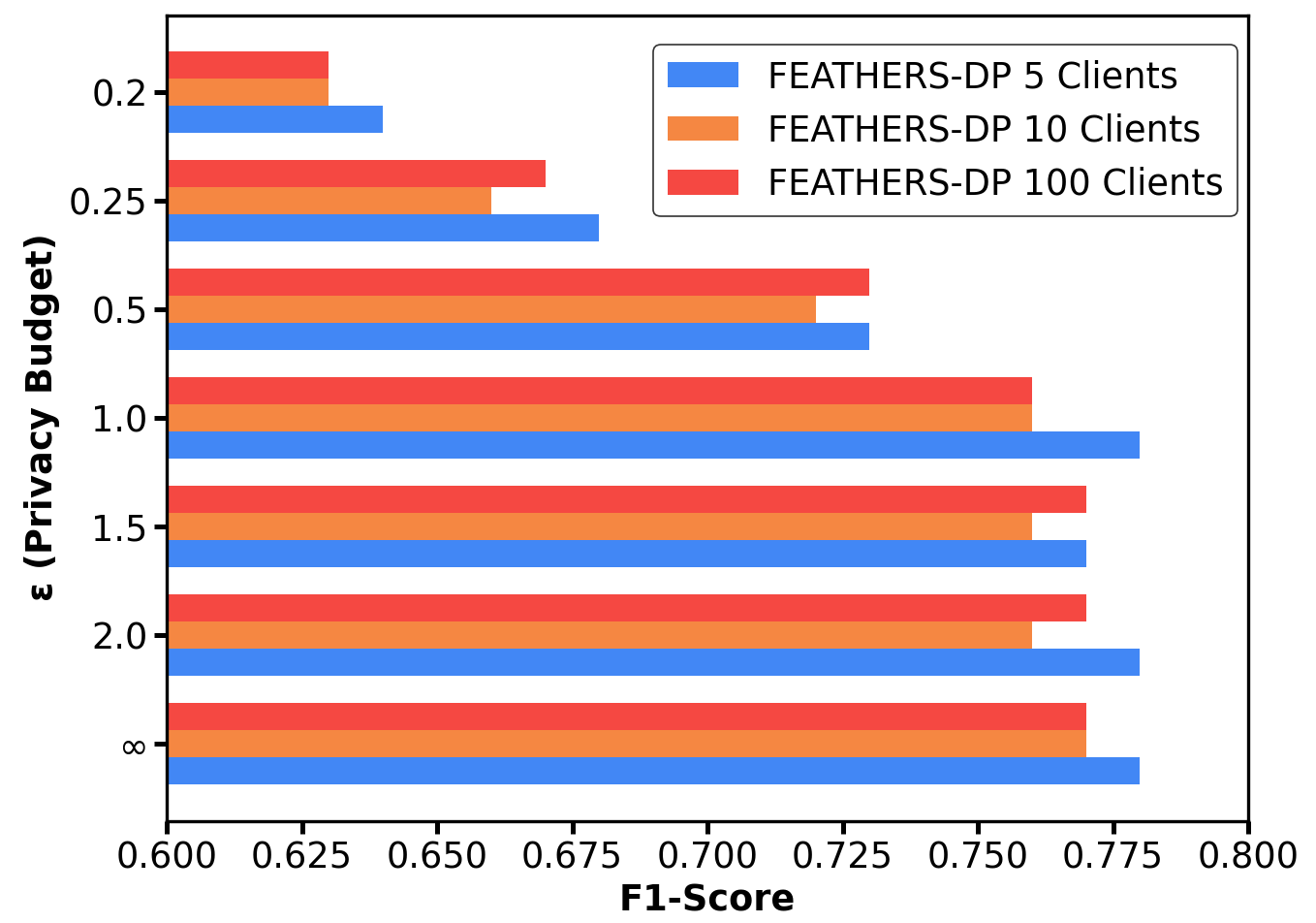}
    \caption{\textbf{DP retains performance.} For $\epsilon \geq 1$, FEATHERS achieves nearly the same performance with DP as without DP regardless of the number of clients. For low $\epsilon$-values (i.e. stronger privacy-guarantees), the performance decreases.}
    \label{fig:f1_vs_epsilon}
\end{figure}

\paragraph{(Q3) FEATHERS preserves privacy.}
To demonstrate that FEATHERS provides privacy guarantees without sacrificing predictive performance, we performed classification on the fraud detection dataset. The hyperparameter search space and the architecture search space are described in Appendix C. The privacy budgets $\epsilon_w, \epsilon_a \in \{ 0.2, 0.25, 0.5, 1.0, 1.5, 2.0, \infty \}$ were both set to equal values in all experiments. It is noteworthy that a privacy budget of $\infty$ corresponds to FEATHERS without DP. The search stage was performed for 100 communication rounds, all other parameters were set as above.
Note that the dataset is heavily skewed (95\% negative class, 5\% positive class). We thus report F1-scores instead of accuracy as high accuracy would not necessarily indicate a good model performance. We further used oversampling on the client-side to account for label-skew. 

Figure \ref{fig:f1_vs_epsilon} visualizes the results for different privacy budgets $\epsilon$. For $\epsilon \geq 1$ we obtained a F1-score of approximately $0.77$. This means, FEATHERS-DP performs equally well as FEATHERS as long as $\epsilon$ is chosen larger to be larger than $1$. Decreasing $\epsilon$ adds more noise on the gradients which increases the privacy level while disturbing the gradient-signal. For $\epsilon \leq 0.5$ we obtained a significant decrease of the F1-score. This confirms that the gradients carrying less (private) information also get less useful for parameter-updates. 

In summary, FEATHERS-DP retains the performance of FEATHERS for appropriate $\epsilon$.
Finally, we emphasize that adding DP came with approximately 1.5-2 times longer runtimes on our setup. A reasonable trade-off to accommodate privacy considerations. 
The underlying reason is that for DP the gradient of each sample has to be manipulated, resulting in poorer parallel execution of automatic differentiation. Hence, two trade-offs have to be considered when using DP: (1) Finding a good balance between a high performing model and privacy guarantees and (2) determining whether a longer runtime for training is still practically feasible to protect privacy. 
\section{Conclusion}
We have introduced FEATHERS, a federated learning method that efficiently optimizes both neural architectures and hyperparameters jointly, while preserving privacy of the underlying training-data. Our empirical investigation demonstrates that FEATHERS is more than competitive with state of the art NAS-algorithms, despite popular approaches like DARTS only performing a subset of the above, while also optimizing in a larger space of hyperparameters.

FEATHERS now allows for a myriad of real-world problems to be addressed fully, e.g. tasks surrounding finance, defence, or healthcare. As one example, hospitals can use it to collaboratively train cancer-detection models without sharing sensible patient-data and without revealing the patient's identity through parameters. That said, we note that fully automating critical systems could be risky and a human in-the-loop should evaluate the quality of the model found before it is deployed and monitor its prediction. 

From a technical perspective, it is further desirable to relax FEATHER's requirement of discrete hyperparameter search spaces in order to properly account for continuous hyperparameters in future work. Incorporating a state reflecting the current state of training in HO-phases to allow for more informed hyperparameter choices instead of a stateless bandit approach is a promising additional direction. Lastly, considering FL-scenarios other than the cross-silo set-up considered in this work is intriguing, also with respect to enabling above mentioned applications. 

\subsubsection*{Acknowledgements}
This work was supported by the ICT-48 Network of AI Research Excellence Center "TAILOR" (EU Horizon 2020, GA No 952215), the Collaboration Lab "AI in Contruction"(AICO), the safeFBDC project and the project National High-Performance Computing for Computational Engineering Sciences (NHR4CES).


\bibliography{hanf}

\begin{thebibliography}{30}
\providecommand{\natexlab}[1]{#1}
\providecommand{\url}[1]{\texttt{#1}}
\expandafter\ifx\csname urlstyle\endcsname\relax
  \providecommand{\doi}[1]{doi: #1}\else
  \providecommand{\doi}{doi: \begingroup \urlstyle{rm}\Url}\fi

\bibitem[Abadi et~al.(2016)Abadi, Chu, Goodfellow, McMahan, Mironov, Talwar,
  and Zhang]{zhan2016DPSGD}
Martin Abadi, Andy Chu, Ian Goodfellow, H.~Brendan McMahan, Ilya Mironov, Kunal
  Talwar, and Li~Zhang.
\newblock Deep learning with differential privacy.
\newblock In \emph{Proceedings of the 2016 ACM SIGSAC Conference on Computer
  and Communications Security (CCS)}, 2016.

\bibitem[Agrawal et~al.(2021)Agrawal, Sarkar, Alazab, Maddikunta, Gadekallu,
  and Pham]{agrawal2021genetic}
Shaashwat Agrawal, Sagnik Sarkar, Mamoun Alazab, Praveen Kumar~Reddy
  Maddikunta, Thippa~Reddy Gadekallu, and Quoc-Viet Pham.
\newblock Genetic cfl: Hyperparameter optimization in clustered federated
  learning.
\newblock \emph{Computational Intelligence and Neuroscience}, 2021.

\bibitem[Dai et~al.(2021)Dai, Low, and Jaillet]{dai2021differentially}
Zhongxiang Dai, Bryan Kian~Hsiang Low, and Patrick Jaillet.
\newblock Differentially private federated bayesian optimization with
  distributed exploration.
\newblock \emph{Advances in Neural Information Processing Systems (NeurIPS)},
  2021.

\bibitem[Darwish et~al.(2020)Darwish, Hassanien, and Das]{darwish2020survey}
Ashraf Darwish, Aboul~Ella Hassanien, and Swagatam Das.
\newblock A survey of swarm and evolutionary computing approaches for deep
  learning.
\newblock \emph{Artificial Intelligence Review}, 2020.

\bibitem[Dong and Yang(2019)]{dong2019searching}
Xuanyi Dong and Yi~Yang.
\newblock Searching for a robust neural architecture in four gpu hours.
\newblock In \emph{Proceedings of the IEEE/CVF Conference on Computer Vision
  and Pattern Recognition}, 2019.

\bibitem[Dwork(2006)]{dwork2006DP}
Cynthia Dwork.
\newblock Differential privacy.
\newblock In Michele Bugliesi, Bart Preneel, Vladimiro Sassone, and Ingo
  Wegener, editors, \emph{Automata, Languages and Programming}, 2006.

\bibitem[Dwork et~al.(2014)Dwork, Roth, et~al.]{dwork2014DP}
Cynthia Dwork, Aaron Roth, et~al.
\newblock The algorithmic foundations of differential privacy.
\newblock \emph{Foundations and Trends in Theoretical Computer Science}, 2014.

\bibitem[Fredrikson et~al.(2015)Fredrikson, Jha, and
  Ristenpart]{fredrikson2015model_inversion}
Matt Fredrikson, Somesh Jha, and Thomas Ristenpart.
\newblock Model inversion attacks that exploit confidence information and basic
  countermeasures.
\newblock In \emph{Proceedings of the 22nd ACM SIGSAC Conference on Computer
  and Communications Security (CCS)}, 2015.

\bibitem[Galv{\'a}n and Mooney(2021)]{galvan2021neuroevolution}
Edgar Galv{\'a}n and Peter Mooney.
\newblock Neuroevolution in deep neural networks: Current trends and future
  challenges.
\newblock \emph{IEEE Transactions on Artificial Intelligence}, 2021.

\bibitem[He et~al.(2020{\natexlab{a}})He, Mushtaq, Ding, and
  Avestimehr]{he2021fednas}
Chaoyang He, Erum Mushtaq, Jie Ding, and Salman Avestimehr.
\newblock Fednas: Federated deep learning via neural architecture search.
\newblock \emph{Workshop on Neural Architecture Search and Beyond for
  Representation Learning (CVPR)}, 2020{\natexlab{a}}.

\bibitem[He et~al.(2020{\natexlab{b}})He, Ye, Shen, and Zhang]{he2020milenas}
Chaoyang He, Haishan Ye, Li~Shen, and Tong Zhang.
\newblock Milenas: Efficient neural architecture search via mixed-level
  reformulation.
\newblock In \emph{Proceedings of the IEEE/CVF Conference on Computer Vision
  and Pattern Recognition}, 2020{\natexlab{b}}.

\bibitem[Kairouz et~al.(2021)Kairouz, McMahan, Avent, Bellet, Bennis, Bhagoji,
  Bonawitz, Charles, Cormode, Cummings, et~al.]{kairouz2021advances}
Peter Kairouz, H~Brendan McMahan, Brendan Avent, Aur{\'e}lien Bellet, Mehdi
  Bennis, Arjun~Nitin Bhagoji, Kallista Bonawitz, Zachary Charles, Graham
  Cormode, Rachel Cummings, et~al.
\newblock Advances and open problems in federated learning.
\newblock \emph{Foundations and Trends in Machine Learning}, 2021.

\bibitem[Khodak et~al.(2021)Khodak, Tu, Li, Li, Balcan, Smith, and
  Talwalkar]{khodak2021federated}
Mikhail Khodak, Renbo Tu, Tian Li, Liam Li, Maria-Florina~F Balcan, Virginia
  Smith, and Ameet Talwalkar.
\newblock Federated hyperparameter tuning: Challenges, baselines, and
  connections to weight-sharing.
\newblock \emph{Advances in Neural Information Processing Systems (NeurIPS)},
  2021.

\bibitem[Koskela and Honkela(2018)]{koskela2018learning}
Antti Koskela and Antti Honkela.
\newblock Learning rate adaptation for federated and differentially private
  learning.
\newblock \emph{arXiv:1809.03832}, 2018.

\bibitem[Li et~al.(2020{\natexlab{a}})Li, Qian, Delgadillo, Muller, Thabet, and
  Ghanem]{li2020sgas}
Guohao Li, Guocheng Qian, Itzel~C Delgadillo, Matthias Muller, Ali Thabet, and
  Bernard Ghanem.
\newblock Sgas: Sequential greedy architecture search.
\newblock In \emph{Proceedings of the IEEE/CVF Conference on Computer Vision
  and Pattern Recognition}, 2020{\natexlab{a}}.

\bibitem[Li et~al.(2020{\natexlab{b}})Li, Huang, Yang, Wang, and
  Zhang]{li2019FedAvgConvergence}
Xiang Li, Kaixuan Huang, Wenhao Yang, Shusen Wang, and Zhihua Zhang.
\newblock On the convergence of fedavg on non-iid data.
\newblock In \emph{8th International Conference on Learning Representations,
  {ICLR} 2020, Addis Ababa, Ethiopia, April 26-30, 2020}, 2020{\natexlab{b}}.

\bibitem[Liu et~al.(2019)Liu, Simonyan, and Yang]{liu2018darts}
Hanxiao Liu, Karen Simonyan, and Yiming Yang.
\newblock Darts: Differentiable architecture search.
\newblock In \emph{International Conference on Learning Representations
  (ICLR)}, 2019.

\bibitem[McMahan et~al.(2017)McMahan, Moore, Ramage, Hampson, and
  y~Arcas]{mcmahan2017communication}
Brendan McMahan, Eider Moore, Daniel Ramage, Seth Hampson, and Blaise~Aguera
  y~Arcas.
\newblock Communication-efficient learning of deep networks from decentralized
  data.
\newblock In \emph{Artificial intelligence and statistics}, 2017.

\bibitem[Mostafa(2019)]{mostafa2019robust}
Hesham Mostafa.
\newblock Robust federated learning through representation matching and
  adaptive hyper-parameters.
\newblock \emph{arXiv:1912.13075}, 2019.

\bibitem[Pham et~al.(2018)Pham, Guan, Zoph, Le, and Dean]{pham2018ENAS}
Hieu Pham, Melody~Y. Guan, Barret Zoph, Quoc~V. Le, and Jeff Dean.
\newblock Efficient neural architecture search via parameter sharing.
\newblock \emph{CoRR}, 2018.

\bibitem[Singh et~al.(2020)Singh, Zhou, Yang, Ding, Lin, and
  Xie]{singh2020differentially}
Ishika Singh, Haoyi Zhou, Kunlin Yang, Meng Ding, Bill Lin, and Pengtao Xie.
\newblock Differentially-private federated neural architecture search.
\newblock \emph{arXiv:2006.10559}, 2020.

\bibitem[Sutton and Barto(2018)]{Sutton1998RL}
Richard~S Sutton and Andrew~G Barto.
\newblock \emph{Reinforcement learning: An introduction}.
\newblock 2018.

\bibitem[Xie and Yuille(2017)]{xie2017genetic}
Lingxi Xie and Alan Yuille.
\newblock Genetic cnn.
\newblock In \emph{Proceedings of the IEEE international conference on computer
  vision}, 2017.

\bibitem[Xie et~al.(2018)Xie, Zheng, Liu, and Lin]{xie2018snas}
Sirui Xie, Hehui Zheng, Chunxiao Liu, and Liang Lin.
\newblock Snas: stochastic neural architecture search.
\newblock \emph{arXiv:1812.09926}, 2018.

\bibitem[Ye et~al.(2022)Ye, Maddi, Murakonda, Bindschaedler, and
  Shokri]{Ye2016MembershipInference}
Jiayuan Ye, Aadyaa Maddi, Sasi~Kumar Murakonda, Vincent Bindschaedler, and Reza
  Shokri.
\newblock Enhanced membership inference attacks against machine learning
  models.
\newblock In \emph{Proceedings of the 2022 ACM SIGSAC Conference on Computer
  and Communications Security (CCS)}, 2022.

\bibitem[Zhou et~al.(2021)Zhou, Ram, Salonidis, Baracaldo, Samulowitz, and
  Ludwig]{zhou2021flora}
Yi~Zhou, Parikshit Ram, Theodoros Salonidis, Nathalie Baracaldo, Horst
  Samulowitz, and Heiko Ludwig.
\newblock Flora: Single-shot hyper-parameter optimization for federated
  learning.
\newblock \emph{arXiv:2112.08524}, 2021.

\bibitem[Zhu et~al.(2021)Zhu, Zhang, and Jin]{zhu2021federated}
Hangyu Zhu, Haoyu Zhang, and Yaochu Jin.
\newblock From federated learning to federated neural architecture search: a
  survey.
\newblock \emph{Complex \& Intelligent Systems}, 2021.

\bibitem[Zoph and Le(2017)]{zoph2016neural}
Barret Zoph and Quoc Le.
\newblock Neural architecture search with reinforcement learning.
\newblock In \emph{International Conference on Learning Representations
  (ICLR)}, 2017.

\bibitem[Zoph et~al.(2016)Zoph, Yuret, May, and Knight]{zoph2016transfer}
Barret Zoph, Deniz Yuret, Jonathan May, and Kevin Knight.
\newblock Transfer learning for low-resource neural machine translation.
\newblock In \emph{Proceedings of the 2016 Conference on Empirical Methods in
  Natural Language Processing (EMNLP)}, 2016.

\bibitem[Zoph et~al.(2018)Zoph, Vasudevan, Shlens, and Le]{zoph2018learning}
Barret Zoph, Vijay Vasudevan, Jonathon Shlens, and Quoc~V Le.
\newblock Learning transferable architectures for scalable image recognition.
\newblock In \emph{Proceedings of the IEEE conference on computer vision and
  pattern recognition (CVPR)}, 2018.

\end{thebibliography}
\clearpage
\appendix
\onecolumn
\section{Algorithms}
Algorithm \ref{algo:client} describes the computation-steps performed by each selected client in a communication round during the Search-stage of FEATHERS. The function $\mathcal{L}$ corresponds to the loss-function and the functions wlr($h$) and alr($h$) retrieve the model-weight-learning rate and the architecture-learning rate from a hyperaparameter-configuration $h$ respectively.
\begin{algorithm}
\caption{FEATHERS Framework Client-side Search stage}
\label{algo:client}
\begin{algorithmic}
    \Require Network parameters $\mathbf{w}$ and architecture $\mathbf{a}$
    \Require Hyperparameter configuration $h$
    \Require Data $\mathbf{X}_{train}$, $\mathbf{X}_{val}$, $\mathbf{y}_{train}$ $\mathbf{y}_{val}$
    \State $l_1 \gets \mathcal{L}_{\mathbf{a}, h}(\mathbf{w}, \mathbf{X}_{val}, \mathbf{y}_{val})$
    \State $\mathbf{w}^* \gets \mathbf{w} - \text{wlr}(h)\nabla_{\mathbf{w}}\mathcal{L}_{\mathbf{a}, h}(\mathbf{w}, \mathbf{X}_{train}, \mathbf{y}_{train})$
    \State $\mathbf{a} \gets \mathbf{a} - \text{alr}(h) \nabla_{\mathbf{a}}\mathcal{L}_{\mathbf{a}, h}(\mathbf{w}^*, \mathbf{X}_{val}, \mathbf{y}_{val})$
    \State $\mathbf{w} \gets \mathbf{w} - \text{wlr}(h)\nabla_{\mathbf{w}}\mathcal{L}_{\mathbf{a}, h}(\mathbf{w}, \mathbf{X}_{train}, \mathbf{y}_{train})$
    \State $l_2 \gets \mathcal{L}_{\mathbf{a}, h}(\mathbf{w}, \mathbf{X}_{val}, \mathbf{y}_{val})$
    \State \Return $l_1$, $l_2$, $\mathbf{w}$, $\mathbf{a}$
 \end{algorithmic}
\end{algorithm}

\begin{algorithm}
\caption{FEATHERS Framework Client-side Search stage with DP}
\label{algo:client_dp}
\begin{algorithmic}
    \Require Network parameters $\mathbf{w}$ and architecture $\mathbf{a}$
    \Require Hyperparameter configuration $h$
    \Require Data $\mathbf{X}_{train}$, $\mathbf{X}_{val}$, $\mathbf{y}_{train}$ $\mathbf{y}_{val}$
    \State $l_1 \gets \mathcal{L}_{\mathbf{a}, h}(\mathbf{w}, \mathbf{X}_{val}, \mathbf{y}_{val})$
    \State $\mathbf{w}^* \gets \mathbf{w} - \text{wlr}(h) \nabla_{\mathbf{w}}\mathcal{L}_{\mathbf{a}, h}(\mathbf{w}, \mathbf{X}_{train}, \mathbf{y}_{train})$
    \State $\mathbf{a} \gets \mathbf{a} - \text{alr}(h) \cdot \frac{1}{B}\sum_{i=1}^B \nabla_{\mathbf{a}} \mathcal{L}_{\mathbf{a}}(\mathbf{w}, \mathbf{X}_{val}^{(i)}, \mathbf{y}_{val}^{(i)}) + \mathcal{N}(0, \sigma_a^2C_a^2\mathbf{I})$
    \State $\mathbf{w} \gets \mathbf{w} - \text{wlr}(h) \cdot \frac{1}{B}\sum_{i=1}^B \nabla_{\mathbf{w}} \mathcal{L}_{\mathbf{a}}(\mathbf{w}, \mathbf{X}_{train}^{(i)}, \mathbf{y}_{train}^{(i)}) + \mathcal{N}(0, \sigma_w^2C_w^2\mathbf{I})$
    \State $l_2 \gets \mathcal{L}_{\mathbf{a}, h}(\mathbf{w}, \mathbf{X}_{val}, \mathbf{y}_{val})$
    \State \Return $l_1$, $l_2$, $\mathbf{w}$, $\mathbf{a}$
 \end{algorithmic}
\end{algorithm}

\begin{algorithm}
\caption{FEATHERS Framework Client-side Evaluation stage}
\label{algo:client_eval}
\begin{algorithmic}
    \Require Network parameters $\mathbf{w}$ and architecture $\mathbf{a}$
    \Require Hyperparameter configuration $h$
    \Require Data $\mathbf{X}_{train}$, $\mathbf{X}_{val}$, $\mathbf{y}_{train}$ $\mathbf{y}_{val}$
    \State $l_1 \gets \mathcal{L}_{\mathbf{a}, h}(\mathbf{w}, \mathbf{X}_{val}, \mathbf{y}_{val})$
    \State $\mathbf{w} \gets \mathbf{w} - \text{wlr}(h)\nabla_{\mathbf{w}}\mathcal{L}_{\mathbf{a}, h}(\mathbf{w}, \mathbf{X}_{train}, \mathbf{y}_{train})$
    \State $l_2 \gets \mathcal{L}_{\mathbf{a}, h}(\mathbf{w}, \mathbf{X}_{val}, \mathbf{y}_{val})$
    \State \Return $l_1$, $l_2$, $\mathbf{w}$, $\mathbf{a}$
 \end{algorithmic}
\end{algorithm}

\begin{algorithm}[H]
\caption{FEATHERS Framework Client-side Evaluation stage with DP}
\label{algo:client_eval_dp}
\begin{algorithmic}
    \Require Network parameters $\mathbf{w}$ and architecture $\mathbf{a}$
    \Require Hyperparameter configuration $h$
    \Require Data $\mathbf{X}_{train}$, $\mathbf{X}_{val}$, $\mathbf{y}_{train}$ $\mathbf{y}_{val}$
    \State $l_1 \gets \mathcal{L}_{\mathbf{a}, h}(\mathbf{w}, \mathbf{X}_{val}, \mathbf{y}_{val})$
    \State $\mathbf{w} \gets \mathbf{w} - \text{wlr}(h)\cdot \frac{1}{B}\sum_{i=1}^B \nabla_{\mathbf{w}} \mathcal{L}_{\mathbf{a}}(\mathbf{w}, \mathbf{X}_{train}^{(i)}, \mathbf{y}_{train}^{(i)}) + \mathcal{N}(0, \sigma_w^2C_w^2\mathbf{I})$
    \State $l_2 \gets \mathcal{L}_{\mathbf{a}, h}(\mathbf{w}, \mathbf{X}_{val}, \mathbf{y}_{val})$
    \State \Return $l_1$, $l_2$, $\mathbf{w}$, $\mathbf{a}$
 \end{algorithmic}
\end{algorithm}

During the Evaluation stage FEATHERS just performs regular network-updates using SGD (or SGD-like variants) under a passed hyperparameter instantiation $h$. This is shown in Algorithm \ref{algo:client_eval}.

\section{Fraud Dataset}
The fraud detection dataset we have used consists of features $\mathbf{X} \in \mathbb{R}^7$ describing financial circumstances of anonymized customers and a binary output $\mathbf{y}$ describing whether a certain instance has high or low credit risk. It consists of $10,000,000$ samples. The labels are strongly skewed towards negative classes ($95\%$ negative class, $5\%$ positive class). Due to this we used oversampling during training (in search and evaluation stage) on client-side. Also, in order to speed up training, we only used a uniformly sampled subset of $1,000,000$ samples instead of the entire. As a preprocessing-step standardization of all features was applied. The dataset used can be found here: \url{https://packages.revolutionanalytics.com/datasets/ccFraud.csv}.

\section{Training Details}
\subsection{General Setup}
We trained the supernet in the search stage for $200$ communication rounds on Fashion-MNIST and $500$ communication rounds on CIFAR-10. After each HO-phase we used $15$ communication rounds to perform NAS, the communication rounds per HO were computed based on the entropy of the softmaxed reward-estimates as described in Section 3. We chose to set $\beta = 4$ to allow a maximum of $16$ communication rounds for HO, assuming a uniform distribution over the $120$ hyperparameter instantiations we use as a search space. Also, we set $\alpha=0.65$ in all experiments. In each communication round, all selected clients perform $5$ epochs of local training. Data was shuffled before distributed on each client using a fixed seed. During training we used gradient clipping with a value of $5$ and chose a batch size of $64$. \\
After the Search stage we selected the architecture leading to the highest accuracy score obtained in the $200$/$500$ communication rounds for Fashion-MNIST/CIFAR-10. This architecture was then used to build and train an evaluation network from scratch as described in Section 3. In the evaluation stage we performed HO again and allowed the training to take place for $1500$ communication rounds where selected clients perform one epoch of training in each communication round. We again computed the number of HO-rounds with $\beta=4$ and set the number of rounds training under a certain hyperparameter instantiation $\mathbf{h}$ to $15$ with $5$ epochs of local training on each client. During training the evaluation network we set gradient clipping to $5$ and chose a batch size of $96$ for Fashion-MNIST/CIFAR-10 and $128$ for Tiny-Imagenet to be comparable to DARTS. \\
\subsection{Search Space Image Classification} \label{sec:search_spaces}
\paragraph{Search Stage.}
Our search space for image classification tasks is divided into an architecture search space $\mathcal{A}$ and a non-architectural hyperparameter-search space $\mathcal{H}$. $\mathcal{A}$ was defined as the space of cells consisting of 7 nodes. The nodes are connected by a mixed operation consisting of the following primitives: Separable and dilated separable convolutions of size $3 \times 3$ and $5 \times 5$, max- and average pooling of size $3 \times 3$, an identity operation as well as a \textit{zero}-operation. We used a stride of one if applicable and used padding. Our convolutional operations are defined using the ReLU-Conv-BN order and we apply separable convolutions twice as in DARTS. We stack 8 cells s.t. the input of each cell is the output of its last two predecessors and every third cell is a reduction cell, the rest are normal cells. Normal cells keep the dimensions of the input whereas reduction cells reduce the input's dimensions. The first cell receives the input twice. The output of each cell is defined as the depth-wise concatenation of the representations of its nodes. In case of fraud detection we are using a search space over MLPs since we deal with tabular data. Here, operations are defined as small MLPs which map the input to a lower or higher dimension and which use either Tanh, Sigmoid or ReLU as activation functions. For a detailed description of $\mathcal{A}$ in the MLP-case, see Appendix C.3. \\
The hyperparameter-space $\mathcal{H}$ consists of candidates for the learning rate used for model- and architecture parameter-updates sampled from a log-uniform distribution from $10^{\exp(\mathcal{U}(-4, 0))}$ and $10^{\exp(\mathcal{U}(-5, -1))}$ respectively. Further we included candidates for weight decay used in updates of both parameter-types sampled from $10^{\exp(\mathcal{U}(-5, -1))}$ and we included candidates for the momentum used in SGD-updates of model parameters with momentum candidates sampled from $\mathcal{U}(0.5, 1)$. We used 120 i.i.d. samples from these distributions to define our discrete hyperparameter search space.

\paragraph{Evaluation Stage.}
Here, our search space consists of non-architectural hyperparameters only. We chose to tune for the learning rate, weight decay, momentum and path-dropout. We chose the first three hyperparameters since these directly influence the behavior of the optimizer, thus of the learning progress. Path-dropout was chosen to avoid over-fitting and to demonstrate that FEATHERS is able to select proper configurations to do so.
In the evaluation on Fashion-MNIST and CIFAR-10 the search space over learning-rate, weight-decay and momentum remained the same as in the search stage. For dropout we allowed values between $0$ and $0.5$. In the evaluation on Tiny-Imagenet we defined the search space as follows: The learning rates were sampled from $10^{\exp(\mathcal{U}(-6, -1))}$, weight-decay was sampled from $10^{\exp(\mathcal{U}(-5, -2))}$, momentum and dropout remained the same as for CIFAR-10 and Fashion-MNIST. We sampled 240 candidates instead of 120. \\
In the MLP-case we replaced path-dropout by regular dropout. We sample 120 i.i.d. samples as above, for (path-)dropout we chose to sample from $\mathcal{U}(0, 0.3)$.

\subsection{Search Space Fraud Detection}
In case of the fraud detection dataset, we defined a search space over MLPs. We define a set of 6 operations, 3 of these retain the input-dimensionality and apply a linear transformation followed by either ReLU, Sigmoid or Tanh. The other 3 operations reduce the dimensionality by a factor of $0.8$. Our search space consists of three cells. Each cell represents a simple MLP-layer consisting of one of the 6 operations defined above. In the search stage, all configurations stayed the same as in the case of training on CIFAR-10/FashionMNIST except for the number of communication rounds which was set to 100. In the evaluation stage the number of communication rounds was set to 200, the rest remained the same.

\section{Technical Details}
We make our code publicly available at: {\footnotesize \url{https://anonymous.4open.science/r/FEATHERS-250B/}}. We have used the \textit{flwr} framework \url{https://flower.dev/} to implement federated learning setting in FEATHERS. All experiments were run on Nvidia DGX clusters with A100-GPUs with 40 GB RAM. One GPU was used for the server and clients were distributed over the remaining GPUs. Of course all clients were isolated from each other during the experiments, the only communication allowed happened between clients and server.

\section{Architectures found by FEATHERS}
\begin{figure}[H]
    \centering
    \includegraphics[scale=0.3]{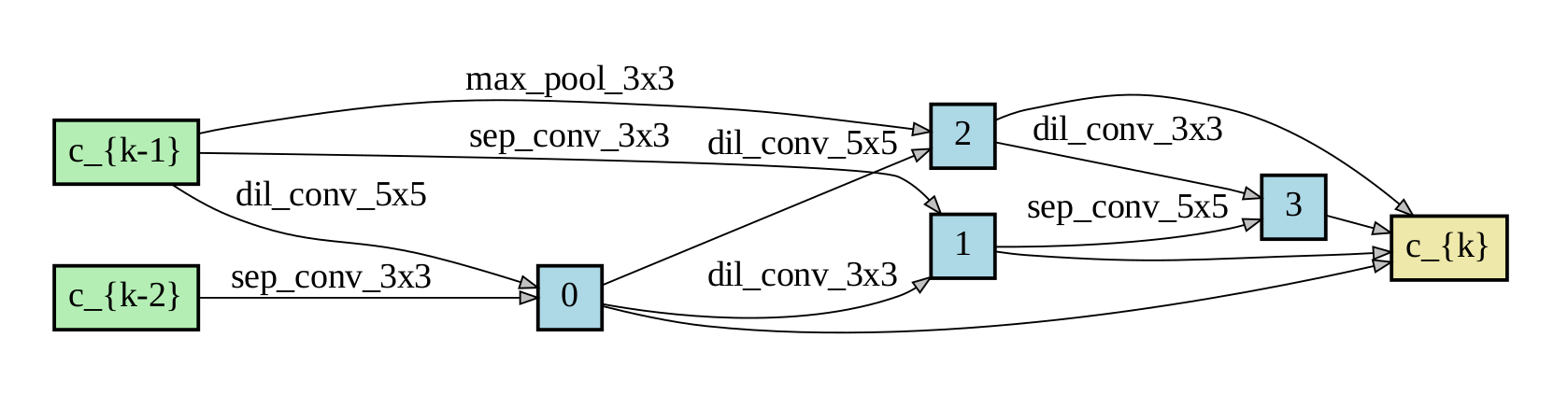}
    \caption{This figure shows the normal cell found by FEATHERS on CIFAR-10, i.e. the dimensions of the input are kept. $c_k$ is the output of the cell, $c_{k - 1}$ and $c_{k -2}$ correspond to the input of the cell (i.e. the output of the last two cells). Each node in the graph shows a representation of the cell-input, computed by the operation(s) shown on the incoming edges of a node.}
    \label{fig:normal_cell}
\end{figure}
\begin{figure}[H]
    \centering
    \includegraphics[scale=0.3]{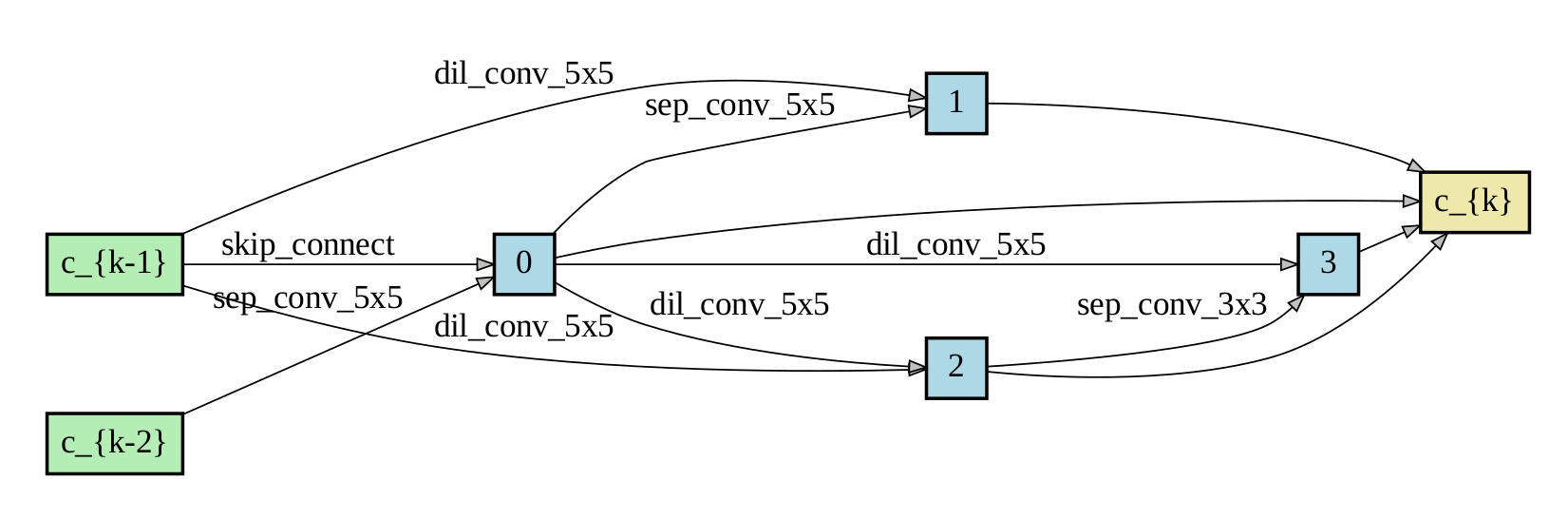}
    \caption{This figure shows the reduction cell found by FEATHERS on CIFAR-10, i.e. the dimensions of the input are reduced. $c_k$ is the output of the cell, $c_{k - 1}$ and $c_{k -2}$ correspond to the input of the cell (i.e. the output of the last two cells). Each node in the graph shows a representation of the cell-input, computed by the operation(s) shown on the incoming edges of a node.}
    \label{fig:reduction_cell}
\end{figure}

\section{Derivation of Selection Rule for \texorpdfstring{$\alpha$}{TEXT}}
Theorem 3 can be used to derive values for $\alpha$ in each HO-round which are guaranteed to be optimal under certain conditions. Based on the assumptions of Theorem 2, we have true mean-values of rewards $\mu^{(j)}$ and $\mu^{(j+1)}$ for round $j$ and $j+1$ respectively. Also, we assume that $|\mu^{(j+1)} - \mu^{(j)}| \leq \delta$ for all $j$. Let's assume we have access to $\mu^{(j)}$ at round $j$ for some $\mathbf{h}$ and that $\mathbf{h}$ is not sampled for $k$ subsequent rounds. Then our estimate of $\mu^{(j+k)}$ would be $\alpha^k \mu^{(j)}$ after $k$ rounds. To obtain the correct estimate of $\mu^{(j+k)}$, the following must hold:
\begin{equation}
    \alpha^k \cdot \mu^{(j)} = \mu^{(j)} + k \cdot (\mu^{(j+1) - \mu^{(j)}})
\end{equation}
We can find the optimal $\alpha$ using the following derivation:
\begin{align}
    \alpha^k \cdot \mu^{(j)} &= \mu^{(j)} + k \cdot (\mu^{(j+1) - \mu^{(j)}}) \\
    \alpha^k &= 1 + \frac{k \cdot (\mu^{(j+1)} - \mu^{(j)})}{\mu^{(j)}} \\
    & \leq 1 + \frac{k \delta}{\mu^{(j)}}
\end{align}
It follows that 
\begin{equation}
    \alpha = \Big(1 + \frac{k \delta}{\mu^{(j)}} \Big)^{\frac{1}{k}}
\end{equation}
Note that in case $\mu^{(j+1)} - \mu^{(j)} = \delta$ by applying the above equation we obtain the optimal $\alpha$.
\newpage
\section{Runtimes}
\begin{table*}[h!]
\centering
\caption{\textbf{FEATHERS does not add significant overhead.} FEATHERS' runtime is approximately $0.4$ GPU-days higher than the runtime of DARTS. FedEx has lower runtimes compared to both, FEATHERS and DARTS, mainly because it does not optimize the architecture and performs less exploration than FEATHERS. All runtimes are reproted in GPU-days.}
\begin{tabular}{c|cccc}
                       & Fsahion-MNIST & CIFAR-10 & Tiny-Imagenet & Fraud Detection \\ \hline
DARTS (5 Clients)      & $1.8$         & $2.1$    & $3.1$         & $0.4$           \\
DARTS (10 Clients)     & $1.7$         & $2.0$    & $2.9$         & $0.3$           \\
DARTS (100 Clients)    & $1.5$         & $1.8$    & $2.7$         & $0.2$           \\
FedEx (5 Clients)      & $0.8$         & $0.9$    & $1.3$         & $0.09$          \\
FedEx (10 Clients)     & $0.8$         & $0.8$    & $1.3$         & $0.07$          \\
FedEx (100 Clients)    & $0.6$         & $0.7$    & $1.1$         & $0.05$          \\
FEATHERS (5 Clients)   & $2.2$         & $2.5$    & $3.6$         & $0.6$           \\
FEATHERS (10 Clients)  & $2.1$         & $2.4$    & $3.5$         & $0.6$           \\
FEATHERS (100 Clients) & $1.9$         & $2.1$    & $3.1$         & $0.35$         
\end{tabular}
\end{table*}

\end{document}